\pdfoutput=1
\documentclass[11pt]{article}

\usepackage[]{acl}

\usepackage{times}
\usepackage{latexsym}

\usepackage{amsmath,amsfonts}
\usepackage{algorithmic}
\usepackage{algorithm}
\usepackage{array}
\usepackage{textcomp}
\usepackage{stfloats}
\usepackage{url}
\usepackage{verbatim}
\usepackage{graphicx}
\usepackage{cite}
\usepackage{multirow}
\usepackage{balance}
\usepackage{threeparttable}
\usepackage{bbm}
\usepackage{bm}
\usepackage{booktabs}
\usepackage{diagbox}
\usepackage{hyperref}

\usepackage[T1]{fontenc}
\usepackage[utf8]{inputenc}

\usepackage{microtype}

\title{Enhancing Cross-lingual Natural Language Inference by Soft Prompting with Multilingual Verbalizer}

\author{Shuang Li$^{1}$, Xuming Hu$^{1}$, Aiwei Liu$^{1}$, Yawen Yang$^{1}$, Fukun Ma$^{1}$, \\ {\bf Philip S. Yu$^{2}$}, {\bf Lijie Wen$^{1*}$}
 \\
  $^{1}$Tsinghua University, $^{2}$University of Illinois Chicago \\
  $^{1}$\texttt{\{lisa18,hxm19,liuaw20,yyw19,mfk22\}@mails.tsinghua.edu.cn}\\
  $^{2}$\texttt{psyu@uic.edu}, $^{1}$\texttt{wenlj@tsinghua.edu.cn}
  }

\begin{document}
\maketitle
\begin{abstract}
Cross-lingual natural language inference is a fundamental problem in cross-lingual language understanding.
Many recent works have used prompt learning to address the lack of annotated parallel corpora in XNLI.
However, these methods adopt discrete prompting by simply translating the templates to the target language and need external expert knowledge to design the templates.
Besides, discrete prompts of human-designed template words are not trainable vectors and can not be migrated to target languages in the inference stage flexibly. In this paper, we propose a novel \textbf{Soft} prompt learning framework with the \textbf{M}ultilingual \textbf{V}erbalizer (SoftMV) for XNLI. 
SoftMV first constructs cloze-style question with soft prompts for the input sample.
Then we leverage bilingual dictionaries to generate an augmented multilingual question for the original question. SoftMV adopts a multilingual verbalizer to align the representations of original and augmented multilingual questions into the same semantic space with consistency regularization.
Experimental results on XNLI demonstrate that SoftMV can achieve state-of-the-art performance and significantly outperform the previous methods under the few-shot and full-shot cross-lingual transfer settings\footnote{The source code will be available at \url{https://github.com/THU-BPM/SoftMV}.\par $\ ^{*}$Corresponding Author.}.
\end{abstract}

\section{Introduction}

Multilingual NLP systems have gained more attention due to the increasing demand for multilingual services. Cross-lingual language understanding (XLU) plays a crucial role in multilingual systems, in which cross-lingual natural language inference (XNLI) is a fundamental and challenging task \citep{conneau-etal-2018-xnli,maccartney-manning-2008-modeling,li2023multi,li2022pair}. NLI is a fundamental problem in NLU that could help with tasks like semantic parsing \citep{liu2022character,lin2022makes}, and relation extraction \citep{liu2022hierarchical, hu-etal-2020-selfore,hu-etal-2021-semi-supervised}. In XNLI settings, the model is trained on the source language with annotated data to reason the relationship between a pair of sentences (namely premise and hypothesis) and evaluated on the target language without parallel corpora.

\begin{table}[htbp]
      \centering
      \resizebox{1\linewidth}{!}{
        \begin{tabular}{ll}
        \toprule
        Type & Prompt Templates \\
        \midrule
        DP & \underline{Premise}. \texttt{Question}: \underline{Hypothesis}? \texttt{Answer}: <MASK>. \\
        SP & \underline{Premise}. \underline{Hypothesis}? <$v_1$>...<$v_n$> <MASK>. \\
        MP & \underline{Premise}. \texttt{Question}: \underline{Hypothesis}? <$v_1$>...<$v_n$> \texttt{Answer}: <MASK>. \\
        \bottomrule
        \end{tabular}%
      }
      \caption{The example of prompt templates for Discrete Prompts (DP), Soft Prompts (SP), and Mixed Prompts (MP). \underline{Premise} and \underline{Hypothesis} are a pair of sentences from the NLI dataset. \texttt{Question} and \texttt{Answer} are template words of discrete prompts. <$v_i$> is the trainable vector of soft prompts.}\label{tab:intro}%
\end{table}%

Pre-trained multilingual language models, such as mBERT \citep{devlin-etal-2019-bert}, XLM \citep{conneau2019cross}, and XLM-R \citep{conneau-etal-2020-unsupervised}, have demonstrated promising performance in cross-lingual transfer learning. These language models learn a shared multilingual embedding space to represent words in parallel sentences. However, these models are trained on a large number of parallel corpora, which are not available in many low-resource languages. The major challenge of XNLI is the lack of annotated data for low-resource languages.

To address this problem, some works explored using prompt learning \citep{brown2020language, schick-schutze-2021-exploiting, shin-etal-2020-autoprompt} when adapting pre-trained language models to downstream tasks in cross-lingual scenarios. Prompt learning reformulates the text classification problem into a masked language modeling (MLM) problem by constructing cloze-style questions with a special token \texttt{<MASK>}. The model is trained to predict the masked word in the cloze-style questions.
As shown in Table \ref{tab:intro}, prompt learning can be divided into three types: Discrete Prompts (DP), Soft Prompts (SP), and Mixed Prompts (MP). 
\citet{zhao-schutze-2021-discrete} investigated the effectiveness of prompt learning in multilingual tasks by simply applying soft, discrete, and mixed prompting with a uniform template in English. 
\citet{qi-etal-2022-enhancing} proposed a discrete prompt learning framework that constructs an augmented sample by randomly sampling a template in another language. By comparing the augmented samples and the original samples in the English template, the model can effectively perceive the correspondence between different languages.
However, discrete prompts of human-designed template words require extensive external expert knowledge and are not flexible enough to adapt to different languages. Therefore, the model can't perform well when transferred from high-resource to low-resource languages.

In this paper, we propose a novel \textbf{Soft} prompt learning framework with the \textbf{M}ultilingual \textbf{V}erbalizer (SoftMV) for XNLI.
First, we construct cloze-style questions for the input samples with soft prompts which consist of trainable vectors.
Second, we apply the code-switched substitution strategy \citep{qin2021cosda} to generate multilingual questions which can be regarded as cross-lingual views for the English questions.
Compared with discrete prompts, soft prompts perform prompting directly in the embedding space of the model and can be easily adapted to any language without human-designed templates. 
Both the original and augmented questions are fed into a pre-trained cross-lingual base model. The classification probability distribution is calculated by predicting the masked token with the multilingual verbalizer to reduce the gap between different languages.
Finally, the two probability distributions are regularized by the Kullback-Leibler divergence (KLD) loss \citep{kullback1951information} to align the representations of original and augmented multilingual questions into the same space.
The entire model is trained with a combined objective of the cross-entropy term for classification accuracy and the KLD term for representation consistency. The well-trained soft prompt vectors will be frozen in the inference stage. 
Experimental results on the XNLI benchmark show that SoftMV outperforms the baseline models by a significant margin under both the few-shot and full-shot settings.

Our contributions can be summarized as follows:
\begin{itemize}
      \item We propose a novel \textbf{Soft} prompt learning framework with a \textbf{M}ultilingual \textbf{V}erbalizer (SoftMV) for XNLI.
      SoftMV leverages bilingual dictionaries to generate augmented multilingual code-switched questions for original questions constructed with soft prompts. 
      \item We adopt the multilingual verbalizer to align the representations of original and augmented questions into the same semantic space with consistency regularization.
      \item We conduct extensive experiments on XNLI and demonstrate that SoftMV can significantly outperform the baseline methods under the few-shot and full-shot cross-lingual transfer settings.
\end{itemize}

\section{Related Work}
Early methods for cross-lingual natural language inference are mainly neural networks, such as \citet{conneau-etal-2018-xnli} and \citet{artetxe-schwenk-2019-massively}.
which encode sentences from different languages into the same embedding space via parallel corpora \citep{hermann-blunsom-2014-multilingual}. 
In recent years, large pre-trained cross-lingual language models have demonstrated promising performance. \citet{devlin-etal-2019-bert} extend the basic language model BERT to multilingual scenarios by pre-trained with multilingual corpora. \citet{conneau2019cross} propose a cross-lingual language model (XLM) which enhances BERT with the translation language modeling (TLM) objective. XLM-R \citep{conneau-etal-2020-unsupervised} is an improvement of XLM by training with more languages and more epochs. 
Although these methods do not rely on parallel corpora, they still have limitations because fine-tuning needs annotation efforts which are prohibitively expensive for low-resource languages.

To tackle this problem, some data augmentation methods have been proposed for XNLI. \citet{ahmad-etal-2021-syntax} propose to augment mBERT with universal language syntax using an auxiliary objective for cross-lingual transfer. 
\citet{dong-etal-2021-data} adopt Reorder Augmentation and Semantic Augmentation to synthesize controllable and much less noisy data for XNLI. 
\citet{bari-etal-2021-uxla} improve cross-lingual generalization by unsupervised sample selection and data augmentation from the unlabeled training examples in the target language. 
\citet{zheng-etal-2021-consistency} propose a cross-lingual finetuning method to better utilize four types of data augmentations based on consistency regularization. 
However, these methods do not perform well under the few-shot settings.

Recently, prompt learning \citep{brown2020language, shin-etal-2020-autoprompt, lester-etal-2021-power, vu-etal-2022-spot, li-liang-2021-prefix, qin-eisner-2021-learning, liu-etal-2022-p} has shown promising results in many NLP tasks under the few-shot setting. The key idea of prompt learning for XNLI is reformulating the text classification problem into a masked language modeling problem by constructing cloze-style questions. 
\citet{su-etal-2022-transferability} propose a novel prompt-based transfer learning approach, which first learns a prompt on one or more source tasks and then uses it to initialize the prompt for a target task.
\citet{wu-shi-2022-adversarial} adopt separate soft prompts to learn embeddings enriched with domain knowledge.
\citet{schick-schutze-2021-exploiting} explore discrete prompt learning to NLI with manually defined templates.
\citet{zhao-schutze-2021-discrete} demonstrate that prompt learning outperforms fine-tuning for few-shot XNLI by simply applying soft, discrete, and mixed prompting with a uniform template in English. 
\citet{qi-etal-2022-enhancing} proposed a discrete prompt learning framework that constructs an augmented sample by randomly sampling a template in another language. 
However, discrete prompts of human-designed template words require extensive external expert knowledge and are not flexible enough to adapt to different languages. 
In our work, we adopt trainable soft prompts to capture correspondence between different languages by comparing the augmented multilingual and original questions. 

\begin{figure*}[ht]
      \centering
      \includegraphics[width=1\textwidth]{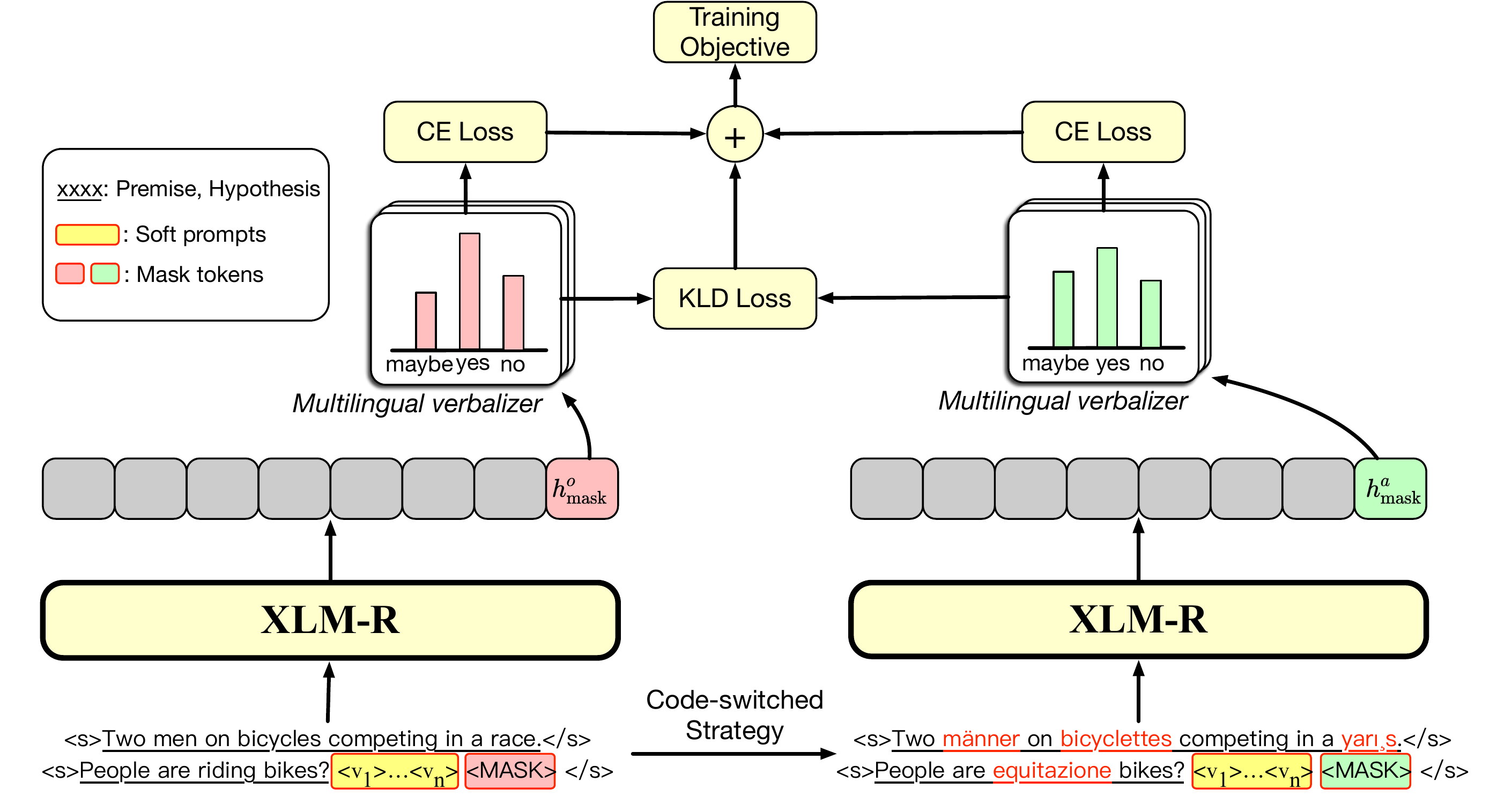} 
      \caption{\label{fig:model}The framework of SoftMV. The left part is the original questions. The right part is the augmented multilingual questions. The model is trained with a combined objective of the cross-entropy losses and the KLD loss.} 
\end{figure*}
\section{Framework}
The proposed SoftMV framework is illustrated in Figure \ref{fig:model}. The training process of SoftMV is formalized in Algorithm \ref{alg1}. For every training triple (premise, hypothesis, label) in English, SoftMV first constructs a cloze-style question with soft prompts initialized from the vocabulary. Then, we apply the code-switched substitution strategy to generate multilingual questions which can be regarded as cross-lingual views for the English questions. Both the original and augmented questions are fed into a pre-trained cross-lingual model to calculate the answer distributions of the mask token with a multilingual verbalizer. SoftMV is trained by minimizing the cross-entropy loss for classification accuracy and the Kullback-Leibler divergence (KLD) loss for representation consistency. Finally, the well-trained soft prompt vectors are frozen in the inference stage.
\subsection{Soft Prompting}
Each instance in batch $\mathcal{I}$ in XNLI dataset is denoted as $(P_i, H_i, Y_i)_{i \in \mathcal{I}}$, where $P_i = \{w_j^{P}\}_{j=1}^m$ denotes the word sequence of premise, $H_i = \{w_j^{H}\}_{j=1}^n$ denotes the word sequence of hypothesis, and $Y_i \in \mathcal{Y} $ denotes the class label. SoftMV first constructs a cloze-style question with soft prompts as illustrated in Table \ref{tab:intro}. The question template is expressed as ``<s>\underline{Premise}.</s> <s>\underline{Hypothesis}? <$v_1$>...<$v_n$> <MASK></s>", where <s> and </s> are special tokens to separate sentences, <MASK> is the mask token, and $v_i$ is associated with a trainable vector (in the PLM's first embedding layer). Soft prompts are tuned in the continuous space and initialized with the average value of embeddings of the PLM's multilingual vocabulary.
\begin{algorithm}
	\renewcommand{\algorithmicrequire}{\textbf{Input:}}
	\caption{The training process of SoftMV.}
	\label{alg1}
	\begin{algorithmic}[1]
            \REQUIRE the number of epochs $E$ and the training set $\mathbb{D}  = \{(P_i, H_i, Y_i)\}_{i=1}^M$.
		\STATE Reform $\mathbb{D}$ to a set of cloze-style questions $\mathbb{Q} = \{(Q_i, Y_i)\}_{i=1}^M$ with soft prompts for each $(P_i, H_i)$ as illustrated in Figure \ref{fig:model}.
		\STATE Extend the set $\mathbb{Q} = \{(Q_i, Q_i^{a}, Y_i)\}_{i=1}^M$ by generating augmented multilingual questions with the code-switched strategy.
            \STATE Divide $\mathbb{Q}$ into a set of batches $\mathbb{B}$.
		\FOR{epoch $e = 1$ to $E$}
                  \STATE Shuffle $\mathbb{B}$.
                  \FOR{each batch $\{(Q_i, Q_i^{a}, Y_i)\}_{1 \leq i \leq N}$ in $\mathbb{B}$}
                        \STATE Compute total loss $\mathcal{L}$ by Eq. \ref{eq:loss}.
                        \STATE Update the parameters $\theta$.
                  \ENDFOR
            \ENDFOR
		
	\end{algorithmic}  
\end{algorithm}
In cross-lingual transfer scenarios, it's a challenge for a model to align contextualized representations in different languages into the same semantic space when trained solely on the English dataset.
Therefore, we adopt the code-switched strategy to create multilingual augmentations for the original questions. Followed by \citet{qin2021cosda}, we use bilingual dictionaries \citep{lample2018word} to replace the words of the original sentences. Specifically, for the English sentence, we randomly choose $n = \alpha*l$ words to be replaced with a translation word from a bilingual dictionary, where $\alpha$ is the code-switched rate and $l$ is the length of the sentence. For example, given the sentence ``Two men on bicycles competing in a race.'' in English, we can generate a multilingual code-switched sample ``Two Männer(DE) on Bicyclettes(FR) competing in a yarış(TR).''  which can be regarded as the cross-lingual view of the same meaning across different languages. 
The original and augmented cloze-style questions are fed into a pre-trained cross-lingual model to obtain the contextualized representation of the mask token, denoted as $h_{\text{mask}}^{o}$ and $h_{\text{mask}}^{a}$. 
Let $l$ denote the size of the vocabulary and $d$ the dimension of the representation of the mask token, the answer probability distribution of the original question is calculated by:

\begin{equation}
      y^o = softmax(\mathbf{W}h_{\text{mask}}^{o}),
\end{equation}

where $\mathbf{W} \in \mathbb{R}^{l \times d}$ is the trainable parameters of the pre-trained MLM layer. The answer probability distribution $y^a$ of the augmented question is calculated in the same way.

\subsection{Multilingual Verbalizer}
After calculating the answer probability distribution of the mask token, we use the verbalizer to calculate the classification probability distribution. The verbalizer $\mathcal{M} \rightarrow \mathcal{V} $ is a function that maps NLI labels to indices of answer words in the given vocabulary. The model is trained to predict masked words that correspond to classification labels, as determined by the verbalizer. Concretely, the verbalizer of English is defined as \{``Entailment'' $\rightarrow$ ``yes''; ``Contradiction'' $\rightarrow$ ``no''; ``Neutral'' $\rightarrow$ ``maybe''\} according to \citet{schick-schutze-2021-just}.

Without parallel corpora in cross-lingual scenarios, there is a gap in the classification space between the original and multilingual representations. Using the English verbalizer for all languages might hinder the model's ability to capture semantic representations for multilingual inputs. Thus we use a multilingual verbalizer to learn a consistent classification probability distribution across different languages. The multilingual verbalizer comprises a set of verbalizers for different languages. The multilingual verbalizer is denoted as $\{\mathcal{M}_l, l \in \mathcal{L} \}$, where $\mathcal{L}$ is the set of languages and $l$ is a specific language. The non-English verbalizers are translated from English using bilingual dictionaries.
Specifically, the verbalizer of Turkish is defined as \{``Entailment'' $\rightarrow$ ``Evet.''; ``Contradiction'' $\rightarrow$ ``hiçbir''; ``Neutral'' $\rightarrow$ ``belki''\}.

\subsection{Training Objective}
In the training stage, given a batch $\mathcal{I}$ of $N$ triples denoted as $(X_i^o, X_i^a, Y_i)_{1\leq i \leq N}$, the cross-entropy losses for the original question $X_i^o$ and the augmented question $X_i^a$ are respectively calculated by:
\begin{align}
      \mathcal{\ell}_{i}^o = -\frac{1}{|\mathcal{L}|} \sum_{l \in \mathcal{L}}\sum_{j=1}^N I(j=\mathcal{M}_l(Y_i)) \operatorname{log} y_{i,j}^o,\\
      \mathcal{\ell}_{i}^a = -\frac{1}{|\mathcal{L}|} \sum_{l \in \mathcal{L}}\sum_{j=1}^N I(j=\mathcal{M}_l(Y_i)) \operatorname{log} y_{i,j}^a,
\end{align}
where $y_{i,j}^o$ (resp. $y_{i,j}^a$) denotes the $j$-th element of the answer probability distribution $y^o$ for the original question $X_i^o$ (resp. for the input $X_i^a$) and $I(C)$ is the indicator function that returns 1 if $C$ is true or 0 otherwise.
The cross-entropy losses of the original and augmented questions on the batch $\mathcal{I}$ are calculated by:
\begin{align}
      \mathcal{L}_{O} = -\frac{1}{N}\sum_{i=1}^N \mathcal{\ell}_{i}^o,\\
      \mathcal{L}_{A} = -\frac{1}{N}\sum_{i=1}^N \mathcal{\ell}_{i}^a.
\end{align}
  
However, for the same premise and hypothesis, the answer probability distribution of the augmented multilingual question created by the code-switched strategy may lead to a deviation from that of the original question due to the misalignment of representations in the multilingual semantic space. Such a deviation may cause the model to learn the wrong probability distribution when the model is evaluated on target languages. 
To alleviate this problem, we propose a consistency regularization to constrain the answer probability distribution. In particular, we adopt the Kullback-Leibler divergence (KLD) to encourage the answer probability distribution of the augmented question to be close to that of the original question. The consistency loss is defined as:
\begin{equation}
      \mathcal{L}_{KLD} = \frac{1}{N}\sum_{i=1}^N (\operatorname{KL}(y_i^o||y_i^a) + \operatorname{KL}(y_i^a||y_i^o)),
\end{equation}

The cross-entropy loss encourages the model to learn correct predictions for the augmented inputs, while the KLD loss enforces consistency between the original and augmented representations in the same multilingual semantic space. Using these loss terms together ensures that the model not only performs well on the original inputs but also generalizes to the augmented inputs, resulting in a more robust model that effectively handles cross-lingual tasks. The overall objective in SoftMV is a tuned linear combination of the cross-entropy losses and KLD loss, defined as:
\begin{equation}
      \mathcal{L} = \lambda_O \mathcal{L}_{O} + \lambda_A \mathcal{L}_{A} + \lambda_{KLD}\mathcal{L}_{KLD},\label{eq:loss}
\end{equation}
where $\lambda_{*}$ are tuning parameters for each loss term. 

\section{Experiment Setup}
\subsection{Benchmark Dataset}
We conducted experiments on the large-scale multilingual benchmark dataset of XNLI \citep{conneau-etal-2018-xnli}, which extends the MultiNLI \citep{williams-etal-2018-broad} benchmark (in English) to 15 languages\footnote{The languages are English (EN), French (FR), Spanish (ES), German (DE), Greek (EL), Bulgarian (BG), Russian (RU), Turkish (TR), Arabic (AR), Vietnamese (VI), Thai (TH), Chinese (ZH), Hindi (HI), Swahili (SW), and Urdu (UR)} through translation and comes with manually annotated development sets and test sets. 
For each language, the training set comprises 393K annotated sentence pairs, whereas the development set and the test set comprise 2.5K and 5K annotated sentence pairs, respectively.

We evaluate SoftMV and other baseline models under the few-shot and full-shot cross-lingual settings, where the models are only trained on English and evaluated on other languages. For the few-shot setting, the training and validation data are sampled by \citet{zhao-schutze-2021-discrete} with $k \in \{1, 2, 4, 8, 16, 32, 64, 128, 256\}$ shots per class from the English training data in XNLI. We report classification accuracy as the evaluation metric.

\subsection{Implementation Details}
We implement SoftMV using the pre-trained XLM-RoBERTa model \citep{conneau-etal-2020-unsupervised} based on PyTorch \citep{paszke2019pytorch} and the Huggingface framework \citep{wolf-etal-2020-transformers}. XLM-R is a widely used multilingual model and the baseline (PCT) we compare with only report the results using XLM-R.

We train our model for 70 epochs with a batch size of 24 using the AdamW optimizer. The hyper-parameter $\alpha$ is set to 0.3 for combining objectives. The maximum sequence length is set to 256. All the experiments are conducted 5 times with different random seeds (\{1, 2, 3, 4, 5\}) and we report the average scores. The trained soft prompt vectors will be frozen in the inference stage.
Appendix \ref{sec:appendix} shows the hyperparameters and computing devices used under different settings in detail.

\subsection{Baseline Models}

We compared SoftMV with the following cross-lingual language models: (1) mBERT \citep{devlin-etal-2019-bert} is a BERT model pre-trained on Wikipedia with 102 languages; (2) XLM \citep{conneau2019cross} is pre-trained for two objectives (MLM and TLM) on Wikipedia with 100 languages; (3) XLM-R \citep{conneau-etal-2020-unsupervised} extends XLM with larger corpora and more epochs; (4) The work \citep{dong-etal-2021-data} proposes an adversarial data augmentation scheme based on XLM-R;
(5) UXLA \citep{bari-etal-2021-uxla} enhances XLM-R with data augmentation and unsupervised sample selection; (6) The work \citep{zhao-schutze-2021-discrete} explores three prompt-learning methods for few-shot XNLI, including DP, SP, and MP; (7) PCT \citep{qi-etal-2022-enhancing} is a discrete prompt learning framework with cross-lingual templates. 

\section{Experiment Results}

\subsection{Main Results}

% Table generated by Excel2LaTeX from sheet 'few-shot'
\begin{table*}[ht!]
      \centering
      \resizebox{1\textwidth}{!}{
      \begin{tabular}{c|l|ccccccccccccccc|c}
      \toprule[1.8pt]
      Shots & Models & EN   & FR   & ES    & DE    & EL    & BG    & RU    & TR    & AR    & VI    & TH    & ZH    & HI    & SW    & UR    & AVG. \\
            \hline
      \multirow{5}[2]{*}{1} 
            & DP    & 33.2  & 34.1  & 33.8  & 33.0  & 33.2  & 33.2  & 33.8  & 34.0  & 32.1  & 32.8  & 33.0  & 33.6  & 33.4  & 33.5  & 32.0  & 33.2 \\
            & SP    & 36.7  & 38.6  & 38.3  & 36.9  & 37.5  & 36.5  & 37.6  & 34.8  & 34.8  & 35.1  & 35.7  & 37.6  & 36.4  & 34.5  & 35.5  & 36.4 \\
            & MP    & 33.3  & 33.7  & 34.0  & 33.0  & 32.1  & 32.3  & 33.0  & 34.6  & 32.3  & 32.8  & 32.2  & 33.4  & 34.1  & 32.9  & 32.7  & 33.1 \\
            & PCT$^\dagger$    & 37.1  & 36.2  & 37.4  & 37.2  & 35.8  & 36.8  & 36.1  & 36.4  & 34.5  & 35.3  & 36.6  & 37.7  & 35.8  & 34.1  & 36.3  & 36.2 \\
            & Ours  & \textbf{43.0} & \textbf{40.1} & \textbf{41.1} & \textbf{39.8} & \textbf{40.2} & \textbf{42.5} & \textbf{44.0} & \textbf{37.4} & \textbf{41.1} & \textbf{41.5} & \textbf{40.4} & \textbf{42.2} & \textbf{40.1} & \textbf{38.3} & \textbf{37.7} & \textbf{40.6} \\
            \hline
      \multirow{5}[2]{*}{2} 
            & DP    & 35.4  & 34.8  & 35.4  & 34.4  & 34.7  & 35.1  & 34.9  & 35.2  & 32.9  & 33.3  & 35.4  & 36.5  & 34.1  & 33.0  & 32.8  & 34.5 \\
            & SP    & 38.0  & 38.6  & 38.2  & 38.2  & 38.4  & 38.1  & 39.2  & 34.8  & 35.9  & 36.7  & 37.2  & 37.7  & 36.3  & 34.4  & 35.5  & 37.1 \\
            & MP    & 34.6  & 34.3  & 33.8  & 34.1  & 33.3  & 34.3  & 34.0  & 34.5  & 32.8  & 33.8  & 34.6  & 35.4  & 33.8  & 33.9  & 32.6  & 34.0 \\
            & PCT$^\dagger$   & 39.3  & 38.4  & 39.0  & 38.7  & 38.9  & 39.2  & 38.8  & 38.2  & 37.6  & 38.1  & 38.4  & 40.1  & 38.2  & 33.7  & 38.0  & 38.3 \\
            & Ours  & \textbf{41.3} & \textbf{42.6} & \textbf{40.9} & \textbf{44.2} & \textbf{42.1} & \textbf{41.7} & \textbf{44.1} & \textbf{40.2} & \textbf{40.2} & \textbf{39.3} & \textbf{40.0} & \textbf{40.8} & \textbf{41.3} & \textbf{37.5} & \textbf{40.4} & \textbf{41.1} \\
            \hline
      \multirow{5}[2]{*}{4} 
            & DP    & 39.5  & 38.3  & 38.9  & 38.9  & 37.7  & 37.6  & 37.5  & 37.2  & 35.4  & 36.0  & 37.8  & 38.7  & 36.4  & 34.7  & 35.9  & 37.4 \\
            & SP    & 41.8  & 41.1  & 39.8  & 40.1  & 40.8  & 40.5  & 41.7  & 35.9  & 38.0  & 37.9  & 39.2  & 39.5  & 37.6  & 35.8  & 37.7  & 39.2 \\
            & MP    & 36.3  & 35.4  & 35.5  & 35.2  & 34.0  & 33.8  & 34.2  & 35.6  & 33.1  & 34.1  & 36.0  & 37.1  & 34.6  & 33.5  & 33.5  & 34.8 \\
            & PCT$^\dagger$    & 41.1  & 39.1  & 40.9  & 41.0  & 39.4  & 39.5  & 40.2  & 39.0  & 37.4  & 38.0  & 38.4  & 40.3  & 37.5  & 35.2  & 37.9  & 39.0 \\
            & Ours  & \textbf{46.8} & \textbf{45.1} & \textbf{45.5} & \textbf{46.4} & \textbf{44.6} & \textbf{44.4} & \textbf{44.8} & \textbf{42.6} & \textbf{40.5} & \textbf{39.6} & \textbf{41.2} & \textbf{43.9} & \textbf{43.3} & \textbf{38.2} & \textbf{42.7} & \textbf{43.3} \\
            \hline
      \multirow{5}[2]{*}{8} 
            & DP    & 36.4  & 35.2  & 35.0  & 34.8  & 34.8  & 34.8  & 34.6  & 34.1  & 32.7  & 33.7  & 35.1  & 35.6  & 33.0  & 32.9  & 33.1  & 34.4 \\
            & SP    & 39.0  & 38.8  & 38.2  & 38.2  & 38.7  & 38.8  & 39.7  & 35.1  & 36.3  & 37.4  & 37.9  & 37.2  & 35.9  & 34.5  & 35.6  & 37.4 \\
            & MP    & 34.8  & 34.8  & 34.7  & 34.8  & 33.2  & 33.2  & 33.8  & 35.1  & 32.7  & 33.6  & 34.5  & 36.3  & 34.8  & 33.1  & 32.7  & 34.1 \\
            & PCT$^\dagger$    & 38.3  & 35.8  & 38.7  & 37.2  & 36.6  & 36.1  & 37.1  & 35.9  & 34.8  & 35.4  & 36.3  & 38.1  & 36.1  & 34.5  & 34.9  & 36.4 \\
            & Ours  & \textbf{47.5} & \textbf{46.7} & \textbf{47.0} & \textbf{46.4} & \textbf{47.5} & \textbf{46.5} & \textbf{46.3} & \textbf{43.7} & \textbf{46.5} & \textbf{45.8} & \textbf{45.1} & \textbf{42.5} & \textbf{43.2} & \textbf{42.1} & \textbf{42.8} & \textbf{45.3} \\
            \hline
      \multirow{5}[2]{*}{16} 
            & DP    & 38.2  & 36.6  & 36.9  & 37.5  & 37.4  & 37.1  & 36.5  & 35.7  & 35.1  & 35.8  & 37.2  & 37.9  & 35.9  & 33.8  & 34.9  & 36.4 \\
            & SP    & 39.5  & 40.9  & 39.4  & 40.2  & 40.4  & 40.6  & 40.6  & 36.3  & 38.9  & 38.5  & 39.5  & 37.4  & 36.9  & 37.1  & 35.9  & 38.8 \\
            & MP    & 33.2  & 34.4  & 34.5  & 34.0  & 32.6  & 33.0  & 33.9  & 34.7  & 32.5  & 33.3  & 33.5  & 35.7  & 34.3  & 33.3  & 32.7  & 33.7 \\
            & PCT   & 46.5  & 44.3  & 41.5  & 36.9  & 45.7  & 40.8  & 42.4  & 43.7  & 43.6  & 44.7  & 43.9  & 44.8  & 44.8  & 40.1  & 42.5  & 43.1 \\
            & Ours  & \textbf{48.8} & \textbf{48.0} & \textbf{47.1} & \textbf{47.7} & \textbf{47.2} & \textbf{47.4} & \textbf{47.8} & \textbf{44.3} & \textbf{45.6} & \textbf{46.6} & \textbf{44.9} & \textbf{46.1} & \textbf{44.9} & \textbf{43.4} & \textbf{43.3} & \textbf{46.2} \\
            \hline
      \multirow{5}[2]{*}{32} 
            & DP    & 43.7  & 43.9  & 42.8  & 43.5  & 42.5  & 43.5  & 42.5  & 42.0  & 41.8  & 41.9  & 40.5  & 39.9  & 39.3  & 37.5  & 39.8  & 41.7 \\
            & SP    & 44.7  & 42.3  & 42.3  & 42.1  & 42.3  & 43.4  & 43.8  & 38.8  & 40.3  & 42.1  & 40.0  & 39.6  & 38.9  & 37.5  & 38.8  & 41.1 \\
            & MP    & 45.5  & 44.7  & 41.2  & 42.6  & 42.3  & 42.2  & 42.2  & 41.2  & 41.0  & 41.7  & 40.2  & 40.9  & 40.2  & 36.5  & 40.5  & 41.5 \\
            & PCT   & 49.6  & 48.8  & 45.5  & 44.4  & 47.4  & 45.4  & 45.5  & 44.3  & 45.7  & 46.7  & 41.6  & 45.6  & 46.7  & 40.3  & 42.9  & 45.4 \\
            & Ours  & \textbf{50.7} & \textbf{48.5} & \textbf{49.1} & \textbf{48.7} & \textbf{48.7} & \textbf{49.8} & \textbf{48.8} & \textbf{47.0} & \textbf{47.9} & \textbf{48.8} & \textbf{45.8} & \textbf{45.1} & \textbf{45.2} & \textbf{43.6} & \textbf{44.9} & \textbf{47.5}\\
            \hline
      \multirow{5}[2]{*}{64} 
            & DP    & 48.9  & 48.0  & 45.0  & 48.1  & 46.9  & 47.6  & 44.9  & 45.7  & 45.6  & 47.3  & 45.7  & 45.2  & 41.6  & 41.0  & 43.3  & 45.7 \\
            & SP    & 49.0  & 46.1  & 45.8  & 46.0  & 43.7  & 43.8  & 44.5  & 41.9  & 43.5  & 45.3  & 44.7  & 44.2  & 40.9  & 40.5  & 40.1  & 44.0 \\
            & MP    & 51.8  & 48.3  & 46.6  & 48.2  & 46.8  & 46.0  & 44.8  & 44.8  & 43.9  & 48.3  & 45.0  & 43.0  & 40.1  & 37.8  & 44.0  & 45.3 \\
            & PCT   & 51.5  & 51.3  & 50.9  & 49.3  & 50.6  & 50.2  & 49.1  & 47.4  & 48.1  & 49.7  & 47.3  & 48.2  & 47.6  & 44.6  & 44.0  & 48.7 \\
            & Ours  & \textbf{54.0} & \textbf{53.6} & \textbf{52.3} & \textbf{51.1} & \textbf{50.7} & \textbf{52.6} & \textbf{51.4} & \textbf{50.1} & \textbf{48.9} & \textbf{51.4} & \textbf{51.2} & \textbf{53.1} & \textbf{51.1} & \textbf{46.3} & \textbf{48.9} & \textbf{51.1} \\
            \hline
      \multirow{5}[2]{*}{128} 
            & DP    & 53.7  & 49.3  & 48.5  & 51.0  & 47.4  & 50.5  & 46.9  & 49.6  & 46.2  & 48.9  & 44.8  & 49.6  & 44.8  & 42.0  & 44.2  & 47.8 \\
            & SP    & 49.5  & 46.4  & 45.8  & 45.0  & 46.3  & 46.2  & 45.0  & 41.9  & 44.8  & 45.0  & 45.6  & 45.7  & 43.3  & 41.2  & 41.2  & 44.9 \\
            & MP    & 52.6  & 50.3  & 49.7  & 49.0  & 49.1  & 48.0  & 46.4  & 48.5  & 46.5  & 48.2  & 48.1  & 50.5  & 47.0  & 42.9  & 44.0  & 48.1 \\
            & PCT   & 55.0  & 53.3  & 53.8  & 52.8  & 53.4  & 51.9  & 51.7  & 50.9  & 50.4  & 51.7  & 50.0  & 51.2  & 51.5  & 47.0  & 47.9  & 51.5 \\
            & Ours  & \textbf{56.6} & \textbf{55.1} & \textbf{55.7} & \textbf{54.7} & \textbf{55.4} & \textbf{55.7} & \textbf{53.7} & \textbf{53.5} & \textbf{52.1} & \textbf{54.5} & \textbf{53.4} & \textbf{54.3} & \textbf{53.1} & \textbf{49.3} & \textbf{51.0} & \textbf{53.9} \\
            \hline
      \multirow{5}[2]{*}{256}
            & DP    & 60.1  & 54.4  & 50.6  & 55.4  & 55.1  & 55.6  & 51.4  & 50.8  & 53.2  & 55.1  & 53.4  & 52.7  & 46.1  & 45.3  & 48.4  & 52.5 \\
            & SP    & 60.6  & 55.8  & 54.8  & 53.0  & 53.1  & 56.0  & 52.5  & 52.1  & 52.3  & 54.5  & 54.5  & 54.6  & 49.4  & 47.3  & 48.5  & 53.3 \\
            & MP    & 60.1  & 55.3  & 51.6  & 50.7  & 54.6  & 54.0  & 53.5  & 51.3  & 52.8  & 52.3  & 53.4  & 53.8  & 49.6  & 45.3  & 47.2  & 52.4 \\
            & PCT   & 60.3  & 58.3  & 58.3  & 56.3  & 57.9  & 56.7  & 55.2  & 54.6  & 54.7  & 57.4  & 55.6  & 55.8  & 54.6  & 51.6  & 52.6  & 56.0 \\
            & Ours  & \textbf{63.3} & \textbf{59.5} & \textbf{61.0} & \textbf{59.5} & \textbf{58.6} & \textbf{60.5} & \textbf{57.8} & \textbf{56.4} & \textbf{58.2} & \textbf{59.2} & \textbf{59.1} & \textbf{60.6} & \textbf{56.1} & \textbf{56.0} & \textbf{53.5} & \textbf{58.6} \\
      \bottomrule[1.8pt]
      \end{tabular}
      }
      \caption{Comparison results on XNLI under the few-shot cross-lingual transfer setting in accuracy(\%). Each number is the mean performance of 5 runs. ``AVG.'' is the average accuracy for 15 languages. PCT$^\dagger$ denote our reproduced results of the model in \citet{qi-etal-2022-enhancing}. The best performance is in \textbf{bold}.}\label{tab:few-shot}
\end{table*}

% Table generated by Excel2LaTeX from sheet 'full'
\begin{table*}[ht!]
      \centering
      \resizebox{1\textwidth}{!}{
        \begin{tabular}{l|ccccccccccccccc|c}
        \toprule
        Models & EN    & FR    & ES    & DE    & EL    & BG    & RU    & TR    & AR    & VI    & TH    & ZH    & HI    & SW    & UR    & AVG. \\
        \midrule
        mBERT & 73.7  & 70.4  & 70.7  & 68.7  & 69.1  & 70.4  & 67.8  & 66.3  & 66.8  & 66.5  & 64.4  & 68.3  & 64.2  & 61.8  & 59.3  & 67.2 \\
        XLM   & 83.2  & 76.7  & 77.7  & 74.0  & 72.7  & 74.1  & 72.7  & 68.7  & 68.6  & 72.9  & 68.9  & 72.5  & 65.6  & 58.2  & 62.4  & 70.7 \\
        XLM-R$_\texttt{base}$ & 84.6  & 78.2  & 79.2  & 77.0  & 75.9  & 77.5  & 75.5  & 72.9  & 72.1  & 74.8  & 71.6  & 73.7  & 69.8  & 64.7  & 65.1  & 74.2 \\
        \citet{dong-etal-2021-data} & 80.8  & 75.8  & 77.3  & 74.5  & 74.9  & 76.3  & 74.9  & 71.4  & 70.0  & 74.5  & 71.6  & 73.6  & 68.5  & 64.8  & 65.7  & 73.0 \\
        DP-XLM-R$_\texttt{base}$ & 83.9  & 78.1  & 78.5  & 76.1  & 75.7  & 77.1  & 75.3  & 73.2  & 71.6  & 74.7  & 70.9  & 73.4  & 70.2  & 63.6  & 65.5  & 73.9 \\
        SP-XLM-R$_\texttt{base}$ & 84.7  & 78.3  & 78.8  & 75.6  & 75.3  & 76.3  & 75.7  & 73.3  & 70.3  & 74.0  & 70.6  & 74.1  & 70.2  & 62.8  & 64.9  & 73.7 \\
        MP-XLM-R$_\texttt{base}$ & 84.2  & 78.4  & 78.8  & 76.9  & 75.3  & 76.5  & 75.7  & 72.7  & 71.2  & 75.2  & 70.8  & 72.8  & 70.7  & 61.5  & 66.0  & 73.8 \\
        PCT-XLM-R$_\texttt{base}$ & 84.9  & 79.4  & 79.7  & 77.7  & 76.6  & 78.9  & 76.9  & 74.0  & 72.9  & 76.0  & 72.0  & 74.9  & 71.7  & 65.9  & 67.3  & 75.3 \\
        SoftMV-XLM-R$_\texttt{base}$ & \textbf{85.2} & \textbf{80.8} & \textbf{79.9} & \textbf{78.7} & \textbf{84.1} & \textbf{81.3} & \textbf{79.5} & \textbf{76.0} & \textbf{77.5} & \textbf{78.8} & \textbf{77.0} & \textbf{76.0} & \textbf{72.0} & \textbf{77.7} & \textbf{77.8} & \textbf{78.8} \\
        \midrule
        XLM-R$_\texttt{large}$ & 88.9  & 83.6  & 84.8  & 83.1  & 82.4  & 83.7  & 80.7  & 79.2  & 79.0  & 80.4  & 77.8  & 79.8  & 76.8  & 72.7  & 73.3  & 80.4 \\
        UXLA & -     & -     & 85.7  & 84.2  & -     & -     & -     & -     & 80.5  & -     & -     & -     & 78.7  & 74.7  & 73.4  & - \\
        PCT-XLM-R$_\texttt{large}$ & 88.3  & 84.2  & 85.1  & 83.7  & 83.1  & 84.4  & 81.9  & 81.2  & 80.9  & 80.7  & 78.8  & 80.3  & 78.4  & 73.6  & 75.6  & 81.3 \\
        SoftMV-XLM-R$_\texttt{large}$ & \textbf{88.9} & \textbf{85.1} & \textbf{85.8} & \textbf{84.2} & \textbf{83.7} & \textbf{85.2} & \textbf{82.3} & \textbf{82.1} & \textbf{81.5} & \textbf{81.4} & \textbf{79.7} & \textbf{81.2} & \textbf{79.1} & \textbf{74.2} & \textbf{76.4} & \textbf{82.1} \\
        \bottomrule
        \end{tabular}%
      }
      \caption{Comparison results on XNLI under the full-shot cross-lingual transfer setting in accuracy(\%). Each number is the mean performance of 5 runs. ``AVG.'' is the average accuracy for 15 languages. The best performance is in \textbf{bold}.}\label{tab:fullshot}%
\end{table*}%

We conducted experiments on the XNLI dataset under the cross-lingual transfer setting, where models are trained on the English dataset and then directly evaluated on the test set of all languages. The settings can be further divided into two sub-settings: the few-shot setting using a fixed number of training samples per class, and the full-shot setting using the whole training set. 

\textbf{Few-shot results} Table \ref{tab:few-shot} reports the results for comparing SoftMV with other models on XNLI under the few-shot setting. 
The results of compared models are taken from \citet{zhao-schutze-2021-discrete, qi-etal-2022-enhancing}. PCT$^\dagger$ in the 1/2/4/8-shot experiments are reproduced by us, for not being reported before.
Note that all models are based on XLM-R$_\texttt{base}$ and trained on the same split of data from \citet{zhao-schutze-2021-discrete}. Results show that SoftMV significantly outperforms all baselines for all languages under all settings by 3.5\% on average. As expected, all models benefit from more shots. When the $k$ shots per class decrease, the gap between the performance of SoftMV and the state-of-the-art model (PCT) becomes larger, implying our model has a stronger ability to align contextualized representations in different languages into the same space when training data are fewer.
In particular, SoftMV outperforms PCT by 4.4\%, 2.8\%, 4.3\%, and 8.9\% in the 1/2/4/8-shot experiments respectively. When the $k$ shots per class are larger than 8, the average performance of SoftMV also outperforms PCT by an absolute gain of 2.5\% on average.
Furthermore, for different languages, all methods perform best on EN (English) and worst on AR (Arabic), VI (Vietnamese), UR (Urdu), and SW (Swahili). It is difficult to obtain usable corpora for these low-resource languages for XLM-R. Thus, the model has a poor learning ability for these languages.
SoftMV also outperforms PCT on these low-resource languages, which demonstrates that our model is more effective in cross-lingual scenarios, especially for low-resource languages.

\textbf{Full-shot results} Table \ref{tab:fullshot} shows the results on XNLI under the full-shot setting. The results of compared models are taken from \citet{qi-etal-2022-enhancing}. SoftMV-XLM-R$_\texttt{base}$ achieves 78.8\% accuracy averaged by 15 target languages, significantly outperforming the basic model XLM-R$_\texttt{base}$ by 4.6\% on average. Compared with PCT, SoftMV improves by 3.5\% on average based on XLM-R$_\texttt{base}$. Furthermore, we can observe that the accuracy of SoftMV exceeds PCT by 0.3\% on EN, but 4.6\% on AR, 11.8\% on SW, and 10.5\% on UR. 
This indicates that SoftMV has better transferability across low-resource languages with well-trained soft prompt vectors.
To further investigate the effectiveness, we also evaluated SoftMV with baselines based on XLM-R$_\texttt{large}$ model.
It can be seen that SoftMV achieves 82.1\% accuracy on average, significantly outperforming PCT and XLM-R$_\texttt{large}$ by 0.8\% and 1.7\%.
Compared with the results on XLM-R$_\texttt{base}$, the improvements of SoftMV on XLM-R$_\texttt{large}$ are smaller, which indicates that SoftMV is more effective on XLM-R$_\texttt{base}$ which has fewer parameters and worse cross-lingual ability. 
The performance gains are due to the stronger ability of SoftMV to align contextualized representations in different languages into the same semantic space with consistency regularization.

% \subsection{Evaluation on Translated Training Data}

\subsection{Ablation Study}

% Table generated by Excel2LaTeX from sheet 'Ablation'
\begin{table*}[ht!]
      \centering
      \resizebox{1\textwidth}{!}{
        \begin{tabular}{l|ccccccccccccccc|c}
        \toprule
        Models & EN    & FR    & ES    & DE    & EL    & BG    & RU    & TR    & AR    & VI    & TH    & ZH    & HI    & SW    & UR    & AVG. \\
        \midrule
        Original & \textbf{47.5} & \textbf{46.7} & \textbf{47.0} & \textbf{46.4} & \textbf{47.5} & \textbf{46.5} & \textbf{46.3} & \textbf{43.7} & \textbf{46.5} & \textbf{45.8} & \textbf{45.1} & \textbf{42.5} & \textbf{43.2} & \textbf{42.1} & \textbf{42.8} & \textbf{45.3} \\
        w/o code-switched & 46.8  & 45.4  & 44.9  & 45.2  & 45.7  & 45.4  & 45.0  & 41.4  & 44.8  & 44.2  & 42.7  & 38.5  & 40.4  & 38.9  & 41.1  & 43.4 \\
        w/o consistency loss & 45.3  & 44.3  & 44.9  & 43.6  & 44.8  & 43.6  & 43.5  & 40.7  & 44.3  & 43.7  & 43.0  & 39.8  & 40.2  & 39.9  & 40.7  & 42.8 \\
        w/o multilingual verbalizer & 44.8  & 44.7  & 44.5  & 43.7  & 45.0  & 44.8  & 44.8  & 43.2  & 43.0  & 43.6  & 43.1  & 42.0  & 42.9  & 41.6  & 42.4  & 43.6 \\
        using discrete prompts & 46.0  & 45.4  & 46.0  & 45.1  & 45.4  & 45.4  & 45.5  & 42.2  & 44.6  & 44.7  & 44.2  & 40.8  & 42.2  & 41.4  & 41.6  & 44.0 \\
        using mixed prompts & 46.2 & 45.8 & 46.1 & 45.6 & 45.7 & 45.1 & 45.8 & 42.3 & 44.7 & 44.9 & 44.6 & 41.0 & 42.5 & 42.0 & 41.7 & 44.3 \\
        using randomly initialized prompts & 47.6  & 46.6  & 46.4  & 45.8  & 46.7  & 45.8  & 44.8  & 43.0  & 46.1  & 45.7  & 44.7  & 42.6  & 42.9  & 40.3  & 42.6  & 44.8 \\
        \bottomrule
        \end{tabular}%
      }
      \caption{Ablation study results for SoftMV under the 8-shot setting in accuracy(\%). ``AVG.'' is the average accuracy for 15 languages.}\label{tab:ablation}%
\end{table*}%
To better understand the contribution of each key component of SoftMV, we conduct an ablation study under the 8-shot setting with XLM-R$_\texttt{base}$. The results are shown in Table \ref{tab:ablation}.
After removing the code-switched method, the performance decreases by 1.9\% on average which shows the augmented multilingual samples can help the model to understand other languages. When we remove the consistency loss, the average accuracy decreases by 2.5\%. The consistency loss can help the model align the representations across different languages into the same semantic space.
Removing the multilingual verbalizer leads to 1.7\% accuracy drop on average. This demonstrates that the multilingual verbalizer can reduce the gap between different languages when calculating the classification probability distribution.
We also replace soft prompts with discrete prompts as illustrated in Table \ref{tab:intro}, which leads to an accuracy drop of 1.3\% on average. The accuracy decreases by 1.0\% when using mixed prompts instead of soft prompts. The reason is that template words in mixed prompts have a bad effect on SoftMV if not specifically designed with expert knowledge.
Furthermore, we use randomly initialized prompts to replace the prompts initialized from the multilingual vocabulary, which leads to 0.5\% accuracy drop on average.

\subsection{Analysis of Code-switched Method}

% Table generated by Excel2LaTeX from sheet 'code-switch'
\begin{figure}[ht]
      \centering
      \includegraphics[width=1\linewidth]{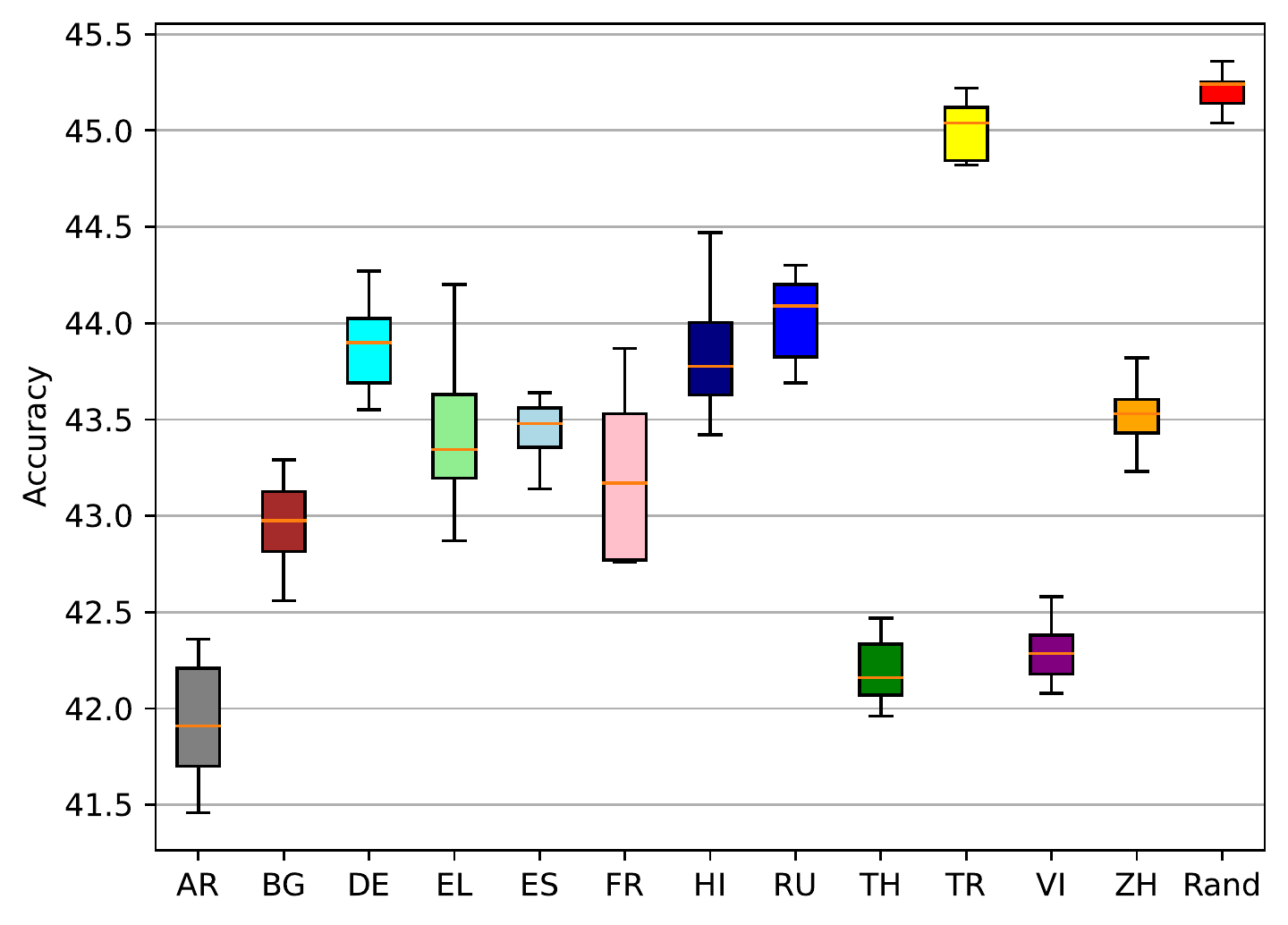} 
      \caption{\label{fig:lang} Evaluation results of different strategies of the code-switched method under the 8-shot setting for 15 languages on average.} 
\end{figure}

To further investigate the code-switched method, we conduct experiments using a single language to create augmented multilingual samples.
Figure \ref{fig:lang} shows the results of SoftMV with 10 different seeds under the 8-shot setting for 15 languages on average. We can observe that SoftMV performs worst with an accuracy of 42.1\% when using AR (Arabic) to replace the words in sentences. When using TR (Turkish) to replace the words in sentences, the performance of SoftMV outperforms the results using another language. The reason is that TR is different from EN, while not too rare like low-resource languages such as UR (Urdu) and AR. Thus the model can better align contextualized representations in different languages into the same semantic space. When randomly selecting languages for the words of each sentence, SoftMV performs best with a lower standard deviation. Therefore, we apply a random strategy for the code-switched method in our experiments.

\subsection{Analysis of Soft Prompts}
\begin{figure}[ht]
      \centering
      \includegraphics[width=1\linewidth]{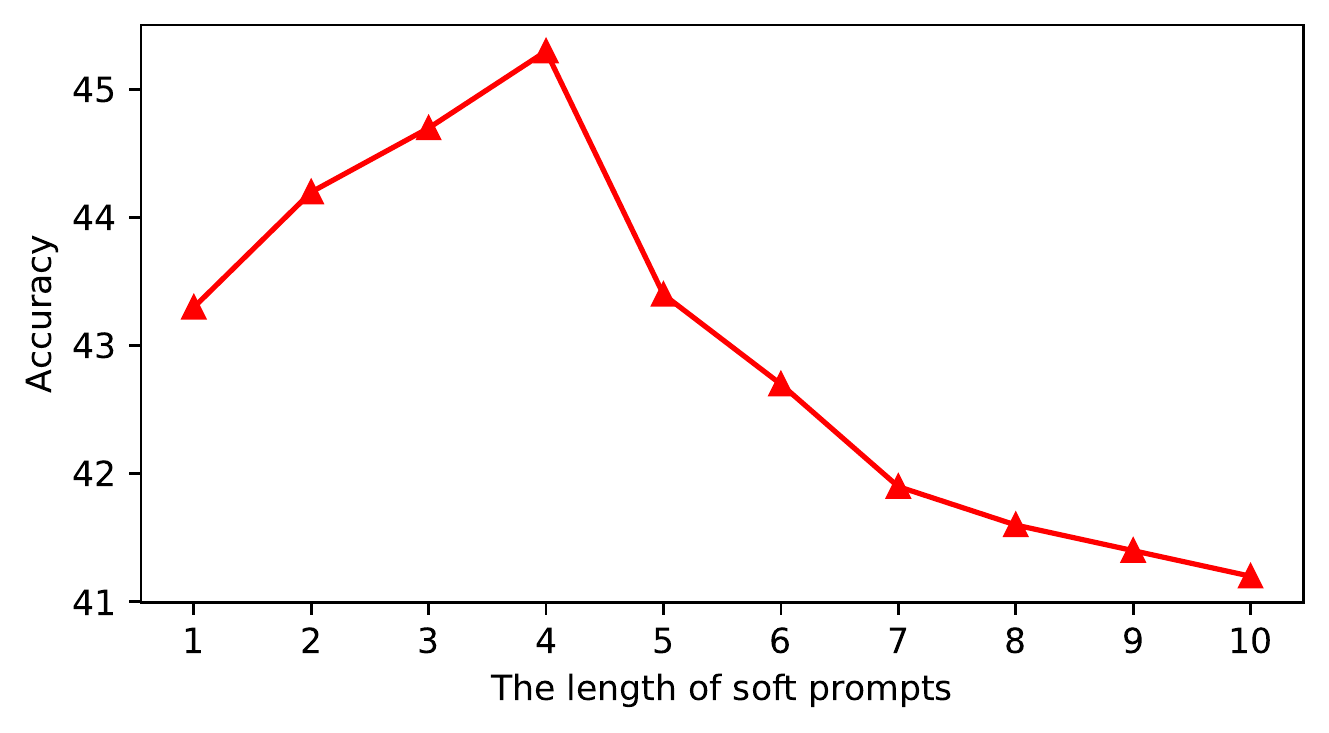} 
      \caption{\label{fig:prompt} Evaluation results of different lengths of soft prompts under the 8-shot setting for 15 languages on average.} 
\end{figure}

We also conducted experiments to show how the length of soft prompts impacts performance.
The results are illustrated in Figure \ref{fig:prompt} under the 8-shot setting. We can observe that the performance of SoftMV is very sensitive to the value of length. As the length of soft prompts increases, the performance of SoftMV first increases and then decreases. As the length of soft prompts increases, the model has the more expressive power to reduce the gaps across different languages.
Therefore, the performance of the model is gradually improved. SoftMV achieves the best performance when the length of soft prompts is 4. When the length is larger than 4, the accuracy decreases sharply. The reason is that the model with longer soft prompts tends to overfit the training data under the few-shot setting.

\section{Conclusion}
In this paper, we propose a novel \textbf{Soft} prompt learning framework with a \textbf{M}ultilingual \textbf{V}erbalizer (SoftMV) for XNLI. SoftMV applies the code-switched substitution strategy to generate multilingual questions for original questions constructed with soft prompts. We adopt the multilingual verbalizer to align the representations of original and augmented samples into the same semantic space with consistency regularization. 
Experimental results on XNLI demonstrate that SoftMV significantly outperforms the previous methods under the few-shot and full-shot cross-lingual transfer settings.
The detailed analysis further confirms the effectiveness of each component in SoftMV. 

\section{Limitations}
SoftMV is specifically designed for cross-lingual natural language inference. We believe that some of the
ideas in our paper can be used in other tasks of XLU, which remains to be further investigated by subsequent research. 

In addition, we conduct experiments on the XNLI dataset which consists of 15 languages. SoftMV outperforms the baseline methods under the cross-lingual transfer settings. However, the cross-lingual ability of SoftMV on other languages, especially those lacking relevant datasets, needs to be verified in future work.

\section*{Acknowledgements}
The work was supported by the National Key Research and Development Program of China (No. 2019YFB1704003), the National Nature Science Foundation of China (No. 62021002), Tsinghua BNRist and Beijing Key Laboratory of Industrial Bigdata System and Application.

\bibliography{anthology,custom}

\begin{thebibliography}{37}
\expandafter\ifx\csname natexlab\endcsname\relax\def\natexlab#1{#1}\fi

\bibitem[{Ahmad et~al.(2021)Ahmad, Li, Chang, and
  Mehdad}]{ahmad-etal-2021-syntax}
Wasi Ahmad, Haoran Li, Kai-Wei Chang, and Yashar Mehdad. 2021.
\newblock \href {https://doi.org/10.18653/v1/2021.acl-long.350}
  {Syntax-augmented multilingual {BERT} for cross-lingual transfer}.
\newblock In \emph{Proceedings of the 59th Annual Meeting of the Association
  for Computational Linguistics and the 11th International Joint Conference on
  Natural Language Processing (Volume 1: Long Papers)}, pages 4538--4554,
  Online. Association for Computational Linguistics.

\bibitem[{Artetxe and Schwenk(2019)}]{artetxe-schwenk-2019-massively}
Mikel Artetxe and Holger Schwenk. 2019.
\newblock \href {https://doi.org/10.1162/tacl_a_00288} {Massively multilingual
  sentence embeddings for zero-shot cross-lingual transfer and beyond}.
\newblock \emph{Transactions of the Association for Computational Linguistics},
  7:597--610.

\bibitem[{Bari et~al.(2021)Bari, Mohiuddin, and Joty}]{bari-etal-2021-uxla}
M~Saiful Bari, Tasnim Mohiuddin, and Shafiq Joty. 2021.
\newblock \href {https://doi.org/10.18653/v1/2021.acl-long.154} {{UXLA}: A
  robust unsupervised data augmentation framework for zero-resource
  cross-lingual {NLP}}.
\newblock In \emph{Proceedings of the 59th Annual Meeting of the Association
  for Computational Linguistics and the 11th International Joint Conference on
  Natural Language Processing (Volume 1: Long Papers)}, pages 1978--1992,
  Online. Association for Computational Linguistics.

\bibitem[{Brown et~al.(2020)Brown, Mann, Ryder, Subbiah, Kaplan, Dhariwal,
  Neelakantan, Shyam, Sastry, Askell et~al.}]{brown2020language}
Tom Brown, Benjamin Mann, Nick Ryder, Melanie Subbiah, Jared~D Kaplan, Prafulla
  Dhariwal, Arvind Neelakantan, Pranav Shyam, Girish Sastry, Amanda Askell,
  et~al. 2020.
\newblock Language models are few-shot learners.
\newblock \emph{Advances in neural information processing systems},
  33:1877--1901.

\bibitem[{Conneau et~al.(2020)Conneau, Khandelwal, Goyal, Chaudhary, Wenzek,
  Guzm{\'a}n, Grave, Ott, Zettlemoyer, and
  Stoyanov}]{conneau-etal-2020-unsupervised}
Alexis Conneau, Kartikay Khandelwal, Naman Goyal, Vishrav Chaudhary, Guillaume
  Wenzek, Francisco Guzm{\'a}n, Edouard Grave, Myle Ott, Luke Zettlemoyer, and
  Veselin Stoyanov. 2020.
\newblock \href {https://doi.org/10.18653/v1/2020.acl-main.747} {Unsupervised
  cross-lingual representation learning at scale}.
\newblock In \emph{Proceedings of the 58th Annual Meeting of the Association
  for Computational Linguistics}, pages 8440--8451, Online. Association for
  Computational Linguistics.

\bibitem[{Conneau and Lample(2019)}]{conneau2019cross}
Alexis Conneau and Guillaume Lample. 2019.
\newblock Cross-lingual language model pretraining.
\newblock \emph{Advances in neural information processing systems}, 32.

\bibitem[{Conneau et~al.(2018)Conneau, Rinott, Lample, Williams, Bowman,
  Schwenk, and Stoyanov}]{conneau-etal-2018-xnli}
Alexis Conneau, Ruty Rinott, Guillaume Lample, Adina Williams, Samuel Bowman,
  Holger Schwenk, and Veselin Stoyanov. 2018.
\newblock \href {https://doi.org/10.18653/v1/D18-1269} {{XNLI}: Evaluating
  cross-lingual sentence representations}.
\newblock In \emph{Proceedings of the 2018 Conference on Empirical Methods in
  Natural Language Processing}, pages 2475--2485, Brussels, Belgium.
  Association for Computational Linguistics.

\bibitem[{Devlin et~al.(2019)Devlin, Chang, Lee, and
  Toutanova}]{devlin-etal-2019-bert}
Jacob Devlin, Ming-Wei Chang, Kenton Lee, and Kristina Toutanova. 2019.
\newblock \href {https://doi.org/10.18653/v1/N19-1423} {{BERT}: Pre-training of
  deep bidirectional transformers for language understanding}.
\newblock In \emph{Proceedings of the 2019 Conference of the North {A}merican
  Chapter of the Association for Computational Linguistics: Human Language
  Technologies, Volume 1 (Long and Short Papers)}, pages 4171--4186,
  Minneapolis, Minnesota. Association for Computational Linguistics.

\bibitem[{Dong et~al.(2021)Dong, Zhu, Fu, Xu, and
  de~Melo}]{dong-etal-2021-data}
Xin Dong, Yaxin Zhu, Zuohui Fu, Dongkuan Xu, and Gerard de~Melo. 2021.
\newblock \href {https://doi.org/10.18653/v1/2021.acl-long.401} {Data
  augmentation with adversarial training for cross-lingual {NLI}}.
\newblock In \emph{Proceedings of the 59th Annual Meeting of the Association
  for Computational Linguistics and the 11th International Joint Conference on
  Natural Language Processing (Volume 1: Long Papers)}, pages 5158--5167,
  Online. Association for Computational Linguistics.

\bibitem[{Hermann and Blunsom(2014)}]{hermann-blunsom-2014-multilingual}
Karl~Moritz Hermann and Phil Blunsom. 2014.
\newblock \href {https://doi.org/10.3115/v1/P14-1006} {Multilingual models for
  compositional distributed semantics}.
\newblock In \emph{Proceedings of the 52nd Annual Meeting of the Association
  for Computational Linguistics (Volume 1: Long Papers)}, pages 58--68,
  Baltimore, Maryland. Association for Computational Linguistics.

\bibitem[{Hu et~al.(2020)Hu, Wen, Xu, Zhang, and Yu}]{hu-etal-2020-selfore}
Xuming Hu, Lijie Wen, Yusong Xu, Chenwei Zhang, and Philip Yu. 2020.
\newblock \href {https://doi.org/10.18653/v1/2020.emnlp-main.299} {{S}elf{ORE}:
  Self-supervised relational feature learning for open relation extraction}.
\newblock In \emph{Proceedings of the 2020 Conference on Empirical Methods in
  Natural Language Processing (EMNLP)}, pages 3673--3682, Online. Association
  for Computational Linguistics.

\bibitem[{Hu et~al.(2021)Hu, Zhang, Ma, Liu, Wen, and
  Yu}]{hu-etal-2021-semi-supervised}
Xuming Hu, Chenwei Zhang, Fukun Ma, Chenyao Liu, Lijie Wen, and Philip~S. Yu.
  2021.
\newblock \href {https://doi.org/10.18653/v1/2021.findings-emnlp.44}
  {Semi-supervised relation extraction via incremental meta self-training}.
\newblock In \emph{Findings of the Association for Computational Linguistics:
  EMNLP 2021}, pages 487--496, Punta Cana, Dominican Republic. Association for
  Computational Linguistics.

\bibitem[{Kullback and Leibler(1951)}]{kullback1951information}
Solomon Kullback and Richard~A Leibler. 1951.
\newblock On information and sufficiency.
\newblock \emph{The annals of mathematical statistics}, 22(1):79--86.

\bibitem[{Lample et~al.(2018)Lample, Conneau, Ranzato, Denoyer, and
  J{\'e}gou}]{lample2018word}
Guillaume Lample, Alexis Conneau, Marc'Aurelio Ranzato, Ludovic Denoyer, and
  Herv{\'e} J{\'e}gou. 2018.
\newblock Word translation without parallel data.
\newblock In \emph{International Conference on Learning Representations}.

\bibitem[{Lester et~al.(2021)Lester, Al-Rfou, and
  Constant}]{lester-etal-2021-power}
Brian Lester, Rami Al-Rfou, and Noah Constant. 2021.
\newblock \href {https://doi.org/10.18653/v1/2021.emnlp-main.243} {The power of
  scale for parameter-efficient prompt tuning}.
\newblock In \emph{Proceedings of the 2021 Conference on Empirical Methods in
  Natural Language Processing}, pages 3045--3059, Online and Punta Cana,
  Dominican Republic. Association for Computational Linguistics.

\bibitem[{Li et~al.(2023)Li, Hu, Lin, Liu, Wen, and Yu}]{li2023multi}
Shuang Li, Xuming Hu, Li~Lin, Aiwei Liu, Lijie Wen, and Philip~S. Yu. 2023.
\newblock \href {https://doi.org/10.1109/TASLP.2023.3270771} {A multi-level
  supervised contrastive learning framework for low-resource natural language
  inference}.
\newblock \emph{IEEE/ACM Transactions on Audio, Speech, and Language
  Processing}, 31:1771--1783.

\bibitem[{Li et~al.(2022)Li, Hu, Lin, and Wen}]{li2022pair}
Shuang Li, Xuming Hu, Li~Lin, and Lijie Wen. 2022.
\newblock Pair-level supervised contrastive learning for natural language
  inference.
\newblock In \emph{Proc. of ICASSP}, pages 8237--8241.

\bibitem[{Li and Liang(2021)}]{li-liang-2021-prefix}
Xiang~Lisa Li and Percy Liang. 2021.
\newblock \href {https://doi.org/10.18653/v1/2021.acl-long.353} {Prefix-tuning:
  Optimizing continuous prompts for generation}.
\newblock In \emph{Proceedings of the 59th Annual Meeting of the Association
  for Computational Linguistics and the 11th International Joint Conference on
  Natural Language Processing (Volume 1: Long Papers)}, pages 4582--4597,
  Online. Association for Computational Linguistics.

\bibitem[{Lin et~al.(2022)Lin, Cao, Huang, Li, Hu, Wen, and
  Wang}]{lin2022makes}
Li~Lin, Yixin Cao, Lifu Huang, Shu'Ang Li, Xuming Hu, Lijie Wen, and Jianmin
  Wang. 2022.
\newblock What makes the story forward? inferring commonsense explanations as
  prompts for future event generation.
\newblock In \emph{Proc. of SIGIR}, pages 1098--1109.

\bibitem[{Liu et~al.(2022{\natexlab{a}})Liu, Yu, Hu, Li, Lin, Ma, Yang, and
  Wen}]{liu2022character}
Aiwei Liu, Honghai Yu, Xuming Hu, Shu'ang Li, Li~Lin, Fukun Ma, Yawen Yang, and
  Lijie Wen. 2022{\natexlab{a}}.
\newblock Character-level white-box adversarial attacks against transformers
  via attachable subwords substitution.
\newblock In \emph{Proc. of EMNLP}.

\bibitem[{Liu et~al.(2022{\natexlab{b}})Liu, Hu, Zhang, Li, Wen, and
  Yu}]{liu2022hierarchical}
Shuliang Liu, Xuming Hu, Chenwei Zhang, Shu'ang Li, Lijie Wen, and Philip~S.
  Yu. 2022{\natexlab{b}}.
\newblock Hiure: Hierarchical exemplar contrastive learning for unsupervised
  relation extraction.
\newblock In \emph{Proc. of NAACL-HLT}, pages 5970--5980.

\bibitem[{Liu et~al.(2022{\natexlab{c}})Liu, Ji, Fu, Tam, Du, Yang, and
  Tang}]{liu-etal-2022-p}
Xiao Liu, Kaixuan Ji, Yicheng Fu, Weng Tam, Zhengxiao Du, Zhilin Yang, and Jie
  Tang. 2022{\natexlab{c}}.
\newblock \href {https://doi.org/10.18653/v1/2022.acl-short.8} {{P}-tuning:
  Prompt tuning can be comparable to fine-tuning across scales and tasks}.
\newblock In \emph{Proceedings of the 60th Annual Meeting of the Association
  for Computational Linguistics (Volume 2: Short Papers)}, pages 61--68,
  Dublin, Ireland. Association for Computational Linguistics.

\bibitem[{MacCartney and Manning(2008)}]{maccartney-manning-2008-modeling}
Bill MacCartney and Christopher~D. Manning. 2008.
\newblock \href {https://aclanthology.org/C08-1066} {Modeling semantic
  containment and exclusion in natural language inference}.
\newblock In \emph{Proceedings of the 22nd International Conference on
  Computational Linguistics (Coling 2008)}, pages 521--528, Manchester, UK.
  Coling 2008 Organizing Committee.

\bibitem[{Paszke et~al.(2019)Paszke, Gross, Massa, Lerer, Bradbury, Chanan,
  Killeen, Lin, Gimelshein, Antiga et~al.}]{paszke2019pytorch}
Adam Paszke, Sam Gross, Francisco Massa, Adam Lerer, James Bradbury, Gregory
  Chanan, Trevor Killeen, Zeming Lin, Natalia Gimelshein, Luca Antiga, et~al.
  2019.
\newblock Pytorch: An imperative style, high-performance deep learning library.
\newblock \emph{Advances in neural information processing systems}, 32.

\bibitem[{Qi et~al.(2022)Qi, Wan, Du, and Chen}]{qi-etal-2022-enhancing}
Kunxun Qi, Hai Wan, Jianfeng Du, and Haolan Chen. 2022.
\newblock \href {https://doi.org/10.18653/v1/2022.acl-long.134} {Enhancing
  cross-lingual natural language inference by prompt-learning from
  cross-lingual templates}.
\newblock In \emph{Proceedings of the 60th Annual Meeting of the Association
  for Computational Linguistics (Volume 1: Long Papers)}, pages 1910--1923,
  Dublin, Ireland. Association for Computational Linguistics.

\bibitem[{Qin and Eisner(2021)}]{qin-eisner-2021-learning}
Guanghui Qin and Jason Eisner. 2021.
\newblock \href {https://doi.org/10.18653/v1/2021.naacl-main.410} {Learning how
  to ask: Querying {LM}s with mixtures of soft prompts}.
\newblock In \emph{Proceedings of the 2021 Conference of the North American
  Chapter of the Association for Computational Linguistics: Human Language
  Technologies}, pages 5203--5212, Online. Association for Computational
  Linguistics.

\bibitem[{Qin et~al.(2021)Qin, Ni, Zhang, and Che}]{qin2021cosda}
Libo Qin, Minheng Ni, Yue Zhang, and Wanxiang Che. 2021.
\newblock Cosda-ml: multi-lingual code-switching data augmentation for
  zero-shot cross-lingual nlp.
\newblock In \emph{Proceedings of the Twenty-Ninth International Conference on
  International Joint Conferences on Artificial Intelligence}, pages
  3853--3860.

\bibitem[{Schick and
  Sch{\"u}tze(2021{\natexlab{a}})}]{schick-schutze-2021-exploiting}
Timo Schick and Hinrich Sch{\"u}tze. 2021{\natexlab{a}}.
\newblock \href {https://doi.org/10.18653/v1/2021.eacl-main.20} {Exploiting
  cloze-questions for few-shot text classification and natural language
  inference}.
\newblock In \emph{Proceedings of the 16th Conference of the European Chapter
  of the Association for Computational Linguistics: Main Volume}, pages
  255--269, Online. Association for Computational Linguistics.

\bibitem[{Schick and
  Sch{\"u}tze(2021{\natexlab{b}})}]{schick-schutze-2021-just}
Timo Schick and Hinrich Sch{\"u}tze. 2021{\natexlab{b}}.
\newblock \href {https://doi.org/10.18653/v1/2021.naacl-main.185} {It{'}s not
  just size that matters: Small language models are also few-shot learners}.
\newblock In \emph{Proceedings of the 2021 Conference of the North American
  Chapter of the Association for Computational Linguistics: Human Language
  Technologies}, pages 2339--2352, Online. Association for Computational
  Linguistics.

\bibitem[{Shin et~al.(2020)Shin, Razeghi, Logan~IV, Wallace, and
  Singh}]{shin-etal-2020-autoprompt}
Taylor Shin, Yasaman Razeghi, Robert~L. Logan~IV, Eric Wallace, and Sameer
  Singh. 2020.
\newblock \href {https://doi.org/10.18653/v1/2020.emnlp-main.346}
  {{A}uto{P}rompt: {E}liciting {K}nowledge from {L}anguage {M}odels with
  {A}utomatically {G}enerated {P}rompts}.
\newblock In \emph{Proceedings of the 2020 Conference on Empirical Methods in
  Natural Language Processing (EMNLP)}, pages 4222--4235, Online. Association
  for Computational Linguistics.

\bibitem[{Su et~al.(2022)Su, Wang, Qin, Chan, Lin, Wang, Wen, Liu, Li, Li, Hou,
  Sun, and Zhou}]{su-etal-2022-transferability}
Yusheng Su, Xiaozhi Wang, Yujia Qin, Chi-Min Chan, Yankai Lin, Huadong Wang,
  Kaiyue Wen, Zhiyuan Liu, Peng Li, Juanzi Li, Lei Hou, Maosong Sun, and Jie
  Zhou. 2022.
\newblock \href {https://doi.org/10.18653/v1/2022.naacl-main.290} {On
  transferability of prompt tuning for natural language processing}.
\newblock In \emph{Proceedings of the 2022 Conference of the North American
  Chapter of the Association for Computational Linguistics: Human Language
  Technologies}, pages 3949--3969, Seattle, United States. Association for
  Computational Linguistics.

\bibitem[{Vu et~al.(2022)Vu, Lester, Constant, Al-Rfou{'}, and
  Cer}]{vu-etal-2022-spot}
Tu~Vu, Brian Lester, Noah Constant, Rami Al-Rfou{'}, and Daniel Cer. 2022.
\newblock \href {https://doi.org/10.18653/v1/2022.acl-long.346} {{SP}o{T}:
  Better frozen model adaptation through soft prompt transfer}.
\newblock In \emph{Proceedings of the 60th Annual Meeting of the Association
  for Computational Linguistics (Volume 1: Long Papers)}, pages 5039--5059,
  Dublin, Ireland. Association for Computational Linguistics.

\bibitem[{Williams et~al.(2018)Williams, Nangia, and
  Bowman}]{williams-etal-2018-broad}
Adina Williams, Nikita Nangia, and Samuel Bowman. 2018.
\newblock \href {https://doi.org/10.18653/v1/N18-1101} {A broad-coverage
  challenge corpus for sentence understanding through inference}.
\newblock In \emph{Proceedings of the 2018 Conference of the North {A}merican
  Chapter of the Association for Computational Linguistics: Human Language
  Technologies, Volume 1 (Long Papers)}, pages 1112--1122, New Orleans,
  Louisiana. Association for Computational Linguistics.

\bibitem[{Wolf et~al.(2020)Wolf, Debut, Sanh, Chaumond, Delangue, Moi, Cistac,
  Rault, Louf, Funtowicz, Davison, Shleifer, von Platen, Ma, Jernite, Plu, Xu,
  Le~Scao, Gugger, Drame, Lhoest, and Rush}]{wolf-etal-2020-transformers}
Thomas Wolf, Lysandre Debut, Victor Sanh, Julien Chaumond, Clement Delangue,
  Anthony Moi, Pierric Cistac, Tim Rault, Remi Louf, Morgan Funtowicz, Joe
  Davison, Sam Shleifer, Patrick von Platen, Clara Ma, Yacine Jernite, Julien
  Plu, Canwen Xu, Teven Le~Scao, Sylvain Gugger, Mariama Drame, Quentin Lhoest,
  and Alexander Rush. 2020.
\newblock \href {https://doi.org/10.18653/v1/2020.emnlp-demos.6} {Transformers:
  State-of-the-art natural language processing}.
\newblock In \emph{Proceedings of the 2020 Conference on Empirical Methods in
  Natural Language Processing: System Demonstrations}, pages 38--45, Online.
  Association for Computational Linguistics.

\bibitem[{Wu and Shi(2022)}]{wu-shi-2022-adversarial}
Hui Wu and Xiaodong Shi. 2022.
\newblock \href {https://doi.org/10.18653/v1/2022.acl-long.174} {Adversarial
  soft prompt tuning for cross-domain sentiment analysis}.
\newblock In \emph{Proceedings of the 60th Annual Meeting of the Association
  for Computational Linguistics (Volume 1: Long Papers)}, pages 2438--2447,
  Dublin, Ireland. Association for Computational Linguistics.

\bibitem[{Zhao and Sch{\"u}tze(2021)}]{zhao-schutze-2021-discrete}
Mengjie Zhao and Hinrich Sch{\"u}tze. 2021.
\newblock \href {https://doi.org/10.18653/v1/2021.emnlp-main.672} {Discrete and
  soft prompting for multilingual models}.
\newblock In \emph{Proceedings of the 2021 Conference on Empirical Methods in
  Natural Language Processing}, pages 8547--8555, Online and Punta Cana,
  Dominican Republic. Association for Computational Linguistics.

\bibitem[{Zheng et~al.(2021)Zheng, Dong, Huang, Wang, Chi, Singhal, Che, Liu,
  Song, and Wei}]{zheng-etal-2021-consistency}
Bo~Zheng, Li~Dong, Shaohan Huang, Wenhui Wang, Zewen Chi, Saksham Singhal,
  Wanxiang Che, Ting Liu, Xia Song, and Furu Wei. 2021.
\newblock \href {https://doi.org/10.18653/v1/2021.acl-long.264} {Consistency
  regularization for cross-lingual fine-tuning}.
\newblock In \emph{Proceedings of the 59th Annual Meeting of the Association
  for Computational Linguistics and the 11th International Joint Conference on
  Natural Language Processing (Volume 1: Long Papers)}, pages 3403--3417,
  Online. Association for Computational Linguistics.

\end{thebibliography}
\bibliographystyle{acl_natbib}

\newpage
\appendix

% Table generated by Excel2LaTeX from sheet 'params'
\begin{table}[ht!]
      \centering
      \resizebox{1\linewidth}{!}{
        \begin{tabular}{cccccc}
        \toprule
        Shots & $\alpha $ & $lr$ & Epochs & Weight decay & Batch size  \\
        \midrule
        1     & 0.10  & 1e-05  & 70 & 0.01 & 12  \\
        2     & 0.10  & 1e-05  & 70 & 0.01 & 12 \\
        4     & 0.10  & 1e-05  & 70 & 0.01 & 12 \\
        8     & 0.15  & 1e-05  & 70 & 0.01 & 12 \\
        16    & 0.20  & 4e-06  & 70 & 0.01 & 12 \\
        32    & 0.15  & 7e-06  & 70 & 0.01 & 12 \\
        64    & 0.15  & 1e-06  & 70 & 0.01 & 12 \\
        128   & 0.20  & 1e-06  & 70 & 0.01 & 12 \\
        256   & 0.35  & 1e-06  & 70 & 0.01 & 12 \\
        Full  & 0.30  & 1e-06  & 70 & 0.01 & 12 \\
        \bottomrule
        \end{tabular}%
      }
      \caption{Hyperparameters used under different settings of XNLI.}\label{tab:params}%
\end{table}%

\section{Training Details}
\label{sec:appendix}

\subsection{Hyperparameters}

Table \ref{tab:params} shows the hyperparameters used under different settings of XNLI. The model is trained for 70 epochs and the checkpoint that performs best on the development set is selected for performance evaluation.

\subsection{Computing Device}
All experiments are conducted on GeForce GTX 3090Ti. We use batch size 24 for a single GPU. Three GPUs are used for few-shot experiments. The full-shot experiments use 6 GPUs.

\vfill
\pagebreak
% \section{Results on Translated Training Data}

\end{document}